\documentclass{article}

 \usepackage[preprint, nonatbib]{neurips_2022}

\usepackage[utf8]{inputenc} 
\usepackage[T1]{fontenc}    
\usepackage{hyperref}       
\usepackage{url}            
\usepackage{booktabs}       
\usepackage{amsfonts}       
\usepackage{nicefrac}       
\usepackage{microtype}      
\usepackage{xcolor}         

\usepackage{pifont}
\usepackage{xcolor}
\usepackage[utf8]{inputenc} 
\usepackage[T1]{fontenc}    
\usepackage{hyperref}       
\usepackage{url}            
\usepackage{booktabs}       
\usepackage{amsfonts}       
\usepackage{nicefrac}       
\usepackage{microtype}      
\usepackage{xcolor}         
\usepackage{comment}
\usepackage{graphicx}
\usepackage{tabularx}
\usepackage{enumitem}
\usepackage{multirow}
\usepackage{amsmath}
\usepackage{wrapfig}
\title{Clinical Contrastive Learning for Biomarker Detection}

\author{%
  Kiran Kokilepersaud, Mohit Prabhushankar, Ghassan AlRegib \\
  Georgia Institute of Technology\\
  Atlanta, GA 30308 \\
  \texttt{\{kpk6,mohit.p,alregib\}@gatech.edu} \\
}

\begin{document}

\vspace{30mm}
\large
\begin{itemize}[leftmargin=2.5cm, align=parleft, labelsep=2cm, itemsep=4ex]

\item[\textbf{Citation}]{K. Kokilepersaud, M. Prabhushankar, and G. AlRegib, "Clinical Contrastive Learning for Biomarker Detection," in NeurIPS 2022 Workshop: Self-Supervised Learning - Theory and Practice, New Orleans, LA, Nov. 28 - Dec. 9 2022.}

\item[\textbf{Review}]{Date of Publication: October 20th, 2022}

\item[\textbf{Code}]{https://github.com/olivesgatech/OLIVES\_Dataset}

\item[\textbf{Bib}]  {@inproceedings\{kokilepersaud2022clinical,\\
    title=\{Clinical Contrastive Learning for Biomarker Detection\},\\
    author=\{Kokilepersaud, Kiran, and Prabhushankar, Mohit and AlRegib, Ghassan\},\\
    booktitle=\{Thirty-sixth Conference on Neural Information Processing Systems\},\\
    year=\{2022\}\}}
\item[\textbf{Contact}]{
\{kpk6, mohit.p, alregib\}@gatech.edu}

\item[\textbf{URL}]{
https://ghassanalregib.info/}

\end{itemize}
\newpage
\clearpage

\maketitle

\begin{abstract}

This paper presents a novel positive and negative set selection strategy for contrastive learning of medical images based on labels that can be extracted from \emph{clinical data}. In the medical field, there exists a variety of labels for data that serve different purposes at different stages of a diagnostic and treatment process. Clinical labels and biomarker labels are two examples. In general, clinical labels are easier to obtain in larger quantities because they are regularly collected during routine clinical care, while biomarker labels require expert analysis and interpretation to obtain. Within the field of ophthalmology, previous work has shown that clinical values exhibit correlations with biomarker structures that manifest within optical coherence tomography (OCT) scans. We exploit  this relationship between clinical and biomarker data to improve performance for biomarker classification. This is accomplished by leveraging the larger amount of clinical data as pseudo-labels for our data without biomarker labels in order to choose positive and negative instances for training a backbone network with a supervised contrastive loss. In this way, a backbone network learns a representation space that aligns with the clinical data distribution available. Afterwards, we fine-tune the network trained in this manner with the smaller amount of biomarker labeled data with a cross-entropy loss in order to classify these key indicators of disease directly from OCT scans. Our method is shown to outperform state of the art self-supervised methods by as much as 5\% in terms of accuracy on individual biomarker detection. 
\end{abstract}

\section*{Introduction}

Contrastive learning \cite{le2020contrastive} refers to a family of self-supervision algorithms that utilize
embedding enforcement
losses with the goal of training a model to learn a rich representation
space. The general premise is that the
model is taught an embedding space where similar pairs of images (positives)
project closer together and dissimilar pairs of images (negatives) project apart. In order to do this in an unsupervised fashion, modern methods, such as \cite{chen2020simple}, generate a positive pair through data augmentation to get multiple views of a sample of interest and treat all other instances within a batch as negatives. While these approaches have shown good performance on the natural image domain, they operate under fundamental assumptions that may not generalize to the medical domain. For example,  data augmentations, by themselves, may not be the best way to choose positive pairs due to small localized regions within medical data oftentimes containing the features most relevant to detect. Additionally, previous approaches operate under the assumption that there exists simply labeled and unlabeled data, rather than distributions of various types of labels as is oftentimes the case within the medical domain. For example, a practical medical task is to detect biomarkers of a disease directly from Optical Coherence Tomography (OCT) scans. This is important because biomarkers refer  
to "any substance, structure, or process that
can be measured in the body or its products and influence or
predict the incidence of outcome or disease \cite{strimbu2010biomarkers}." However, biomarkers are difficult to label in large quantities \cite{mcdonald2015effects} due to the requirement of expert interpretation and analysis. Another reason for this difficulty is that biomarkers such as Diabetic Macular Edema (DME), Intraretinal Fluid (IRF), and others found in Appendix \ref{app:dataset} exist as fine-grained structures that can be difficult to distinguish from the surrounding context. While it is challenging to obtain detailed biomarker information, other types of data are collected more easily as part of standard clinical practice such as Best Central Visual Acutity (BCVA) and Central Subfield Thickness (CST). This information is referred to as clinical labels. 

\begin{figure}[t!]
\centering
\includegraphics[scale = .35]{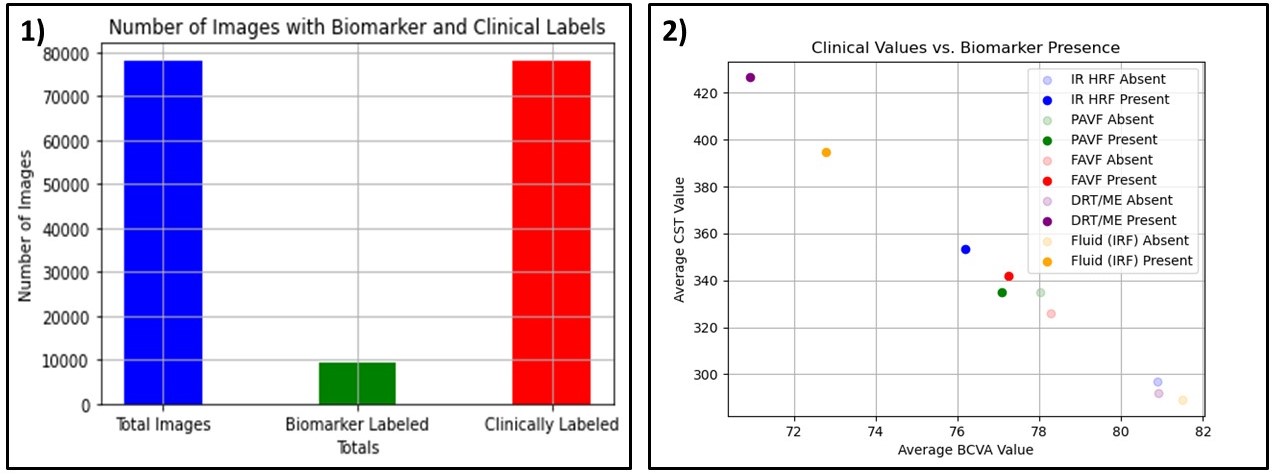}
\caption{This gives an overview of statistics regarding biomarkers and clinical labels. 1) This shows the number of images with biomarker and clinical labels in the \texttt{OLIVES} dataset. 2) All 9408 OCT scans with biomarker labels were grouped based on the presence or absence of a specific biomarker. These biomarker groups were then averaged based on their associated CST and BCVA values. It can be observed that, on average, images with a biomarker present are separable from images with a biomarker absent, with respect to clinical values, thus indicating a relationship between clinical values and biomarkers. }\label{fig:separability}
\end{figure}

To illustrate this point, we show statistics of available data from the \texttt{OLIVES} dataset for ophthalmology  \cite{prabhushankar2022olives} in Figure \ref{fig:separability}. It can be observed that of the 78108 Optical Coherence Tomography (OCT) scans within this dataset, all are labeled with some type of clinical information while a small amount is labeled with biomarker information. Additionally, Figure \ref{fig:separability} shows the relationship between biomarkers and clinical labels. It can be observed that the corresponding clinical values of BCVA and CST are separable depending on whether a specific biomarker is present or absent. Furthermore, work within the medical field \cite{hannouche2012correlation,sun2014disorganization,murakami2011association,kashani2010retinal} has shown that clinical labels can act as indicators of structural changes that manifest within OCT scans. All of this indicates that these clinical labels exhibit non-trivial relationships  with key biomarkers of disease and can potentially act as surrogate labels for biomarkers. We exploit this relationship  by utilizing clinical labels as a means to choose positive and negative sets for a supervised contrastive loss. The representation learnt from training in this manner is then used  to train a linear classifier utilizing a much smaller
subset of biomarker labels.  As a result, the model will be able to
utilize the larger pool of clinically labeled data in order to better learn how to classify specific biomarkers. Contributions of this work include ($i$) introducing a novel contrastive learning strategy based on OCT clinical data, ($ii$) comprehensive analysis of the effect of different medical label distributions on performance, and ($iii$) comparison against state of the art self-supervision algorithms. 

\section*{Theoretical Interpretation}

In \cite{arora2019theoretical} the authors present a theoretical framework for contrastive learning. Let $X$ denote the set of all possible data points. In this framework, contrastive learning assumes access to similar data in the form of $(x,x^{+})$ that comes from a distribution $D_{sim}$ as well as $k$ iid negative samples $x^{-}_{1}, x^{-}_{2}, ..., x^{-}_{k}$ from a distribution $D_{neg}$. This idea of similarity is formalized through the introduction of a set of latent classes $C$ and an associated probability distribution $D_{c}$ over $X$ for every class $c\in C$. $D_{c}(x)$ quantifies how relevant $x$ is to class $c$ with a higher probability assigned to data points belonging to this class. Additionally, let us define $\rho$ as a distribution that describes how these classes naturally occur within the unlabeled data. From this, the positive and negative distribution are characterized as $D_{sim} = \displaystyle \mathop{\mathbb{E}}_{c \sim \rho} D_{c}(x)D_{c}(x^+)$ and $D_{neg} = \displaystyle \mathop{\mathbb{E}}_{c \sim \rho} D_{c}(x^-)$ where $D_{neg}$ is from the marginal of $D_{sim}$. 


The key idea that separates our work from the standard contrastive learning formulation presented above is a deeper look at the relationships between $\rho$, $D_{sim}$, and $D_{neg}$. In principal, during unsupervised training, there is no information that provides the true class distribution $\rho$ of the dataset $X$. The central goal of contrastive learning is to generate an effective $D_{sim}$ and $D_{neg}$ such that the model is guided towards learning $\rho$ by identifying the distinguishing features between the two distributions. Ideally, this guidance occurs through the set of positives belonging to the same class $c_{p}$ and all negatives belonging to any class $c_{n} \neq c_{p}$ as shown in the supervised framework \cite{khosla2020supervised}. Traditional approaches such as \cite{chen2020simple,chen2020improved,li2020prototypical}, enforces positive pair similarity through augmenting a sample to define a positive pair which would clearly represent an instance belonging to the same class. However, these strategies do not define a process by which negative samples are guaranteed to belong to different classes. This problem is discussed in \cite{arora2019theoretical} where the authors decompose the contrastive loss $L_{un}$ as a function of an instance of a hypothesis class $f \in F$ into $L_{un}(f) = (1-\tau) L_{\neq}(f) + (\tau) L_{=}(f)$. This states that the contrastive loss is the sum of the loss suffered when the negative and positive pair come from different classes ($L_{\neq}(f)$) as well as the loss when they come from the same class ($L_{=}(f)$). In an ideal setting ($L_{=}(f)$) would approach 0, but this is impossible without direct access to the underlying class distribution $\rho$. However, it may be the case that there exists another modality of data during training that provides us with a distribution $\rho_{clin}$ with the property that the $KL(\rho_{clin}||\rho) \leq \epsilon$, where $\epsilon$ is sufficiently small. In this case, the $D_{sim}$ and $D_{neg}$ could be drawn from $\rho_{clin}$ in the form:  $D_{sim} = \displaystyle \mathop{\mathbb{E}}_{c \sim \rho_{clin}} D_{c}(x)D_{c}(x^+)$ and $D_{neg} = \displaystyle \mathop{\mathbb{E}}_{c \sim \rho_{clin}} D_{c}(x^-)$. If $\rho_{clin}$ is a sufficiently good approximation for $\rho$, then this has a higher chance for the contrastive loss to choose positives and negatives from different class distributions and have an overall lower resultant loss. In this work, this related distribution that is in excess comes from the availability of clinical information within the unlabeled data and acts to form the $\rho_{clin}$ that we can use for choosing positives and negatives. As discussed in the introduction, this clinical data acts as a surrogate for the true distribution $\rho$ that is based on the severity of disease within the dataset and thus has the theoretical properties discussed.
\vspace{-.6cm}
\section*{Methodology}
 The \texttt{OLIVES} dataset \cite{prabhushankar2022olives} has 9408 OCT scans with explicit biomarker labels for 16 different biomarkers. The biomarker labels are organized as a 16x1 vector for each OCT scan where each entry in the vector is 1 or 0 to indicate the presence or absence of the corresponding biomarker. Of these 16 biomarkers, 5 exist in sufficiently balanced quantities to train a classifier for detection. These are Intra-Retinal Hyper-Reflective Foci (IRHRF), Partially At-
tached Vitreous Face (PAVF), Fully Attached Vitreous Face (FAVF),
Intra-Retinal Fluid (IRF), and Diffuse Retinal Thickening or Macular Edema (DRT/ME). This data also has 78108 OCT Scans with the clinical labels: BCVA, CST, and Eye ID. The clinical labels are discrete measurement values taken at every visit for each individual eye. These images are derived from 96 unique eyes. The dataset is split into a training set of the images from 76 eyes and the test set constitutes the images from the remaining 20 eyes. From these
test OCT scans, random sampling was employed to develop an individualized test set for each of the 5 biomarkers used in our analysis.
This results in a balanced test set, for each biomarker, where 500
OCT Scans have the biomarker present and 500 OCT Scans have the
biomarker absent. Further details on the dataset can be found in Appendix \ref{app:dataset}.

\begin{figure}[t!]
\centering
\includegraphics[scale = .3]{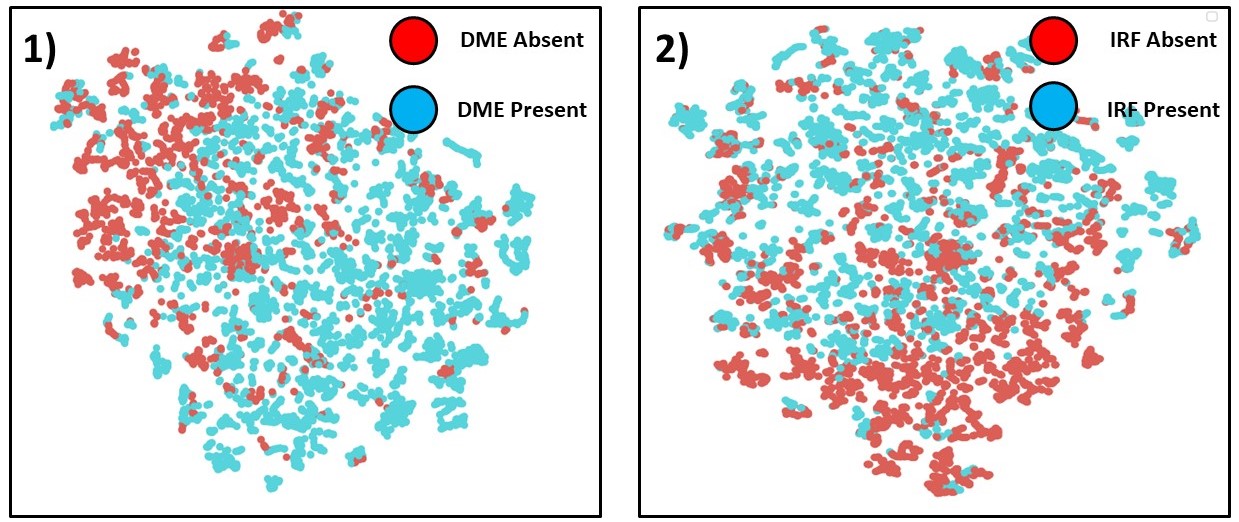}
\caption{T-SNE visualization of the OLIVES Biomarker test set labeled by the presence or absence
of DME and IRF. We can effectively achieve an embedding space that is separable with respect to
biomarkers 1) DME and 2) IRF. }\label{fig:tsne}
\end{figure}

The overall block diagram of the proposed method is summarized in Appendix \ref{app:contrastive} Figure \ref{fig:methodology}. Within the \texttt{OLIVES} dataset, each individual image is associated with the clinical values BCVA, CST, and Eye ID that are taken during the original patient visit. For each experiment, we first choose one of these clinical values to act as a label for each image in the dataset. A backbone network ResNet-18 \cite{he2016deep} $f(.)$ is trained with a supervised clinical contrastive loss \cite{khosla2020supervised} that uses the clinical label to choose positives and negatives as shown in  the first part of Figure \ref{fig:methodology}. After training in this manner, the weights of the backbone network are frozen and a linear layer is appended to the output of this network. This layer is fine-tuned using the smaller subset of images containing explicit biomarker labels as shown in the second part of Figure \ref{fig:methodology}. We observe in Figure \ref{fig:tsne} that a backbone trained with a clinical contrastive loss produces an embedding space that is separable with respect to individual biomarkers. Further details surrounding the training and loss function can be found in Appendix \ref{app:contrastive}.

\vspace{-.6cm}
\section*{Results}
During the clinical contrastive learning step, a single clinical label is chosen to act as the label on which positives and negatives are selected. For example, in Table \ref{tab:main_table}, when CST is specified as the method, this indicates a supervised contrastive loss $L_{CST}$ where CST is utilized as the label of interest for the images in the dataset. The appended linear layer is then fine-tuned by training each biomarker individually as well as in a multi-label setting where all biomarkers are predicted simultaneously. This is reflected in Table \ref{tab:main_table} where performance in predicting each of the 5 biomarkers individually is shown by the Accuracy and F1-score for each biomarker as well as the average AUROC, Specificity , and Sensitivity to get an overall sense of performance across predicting all 5 biomarkers. We compare against 3 state of the art contrastive learning algorithms (SimCLR\cite{chen2020simple}, Moco v2 \cite{chen2020improved}, and PCL \cite{li2020prototypical}) that are trained in the same manner with their own method for choosing positives and negatives. We observe a consistent improvement on these strategies for both individual biomarker classification performance and performance in a multi-label setting. Part of the reason for this improvement may be due to the finer granularity of tasks within the medical domain. Previous work \cite{cole2021does} has shown that standard contrastive learning strategies are worse in situations that involve fine-grained recognition. In this case, we are attempting to detect small perturbations, rather than situations more consistent with the natural image domain where the subject of interest is present throughout the image. This acts as a preliminary explanation why the traditional strategies do comparably well for IRF and DME, but perform much worse for IRHRF, PAVF, and FAVF. IRF and DME are much easier to distinguish from the surroundng context, which is of benefit to these algorithms. This can be understood by the images in Figure \ref{fig:biomarkers}. By leveraging a wider array of positives from the distribution of clinical labels, we observe superior performance on this biomarker detection task.
\begin{table}[t!]
\centering
\caption{Benchmark of the performance of supervised contrastive training on images with clinical and biomarker data. Standard deviations for the results can be found in Table \ref{tab:main_std} of Appendix \ref{app:results}}
\label{tab:main_table}
\resizebox{\textwidth}{!}{%
\begin{tabular}{@{}cccccccccccccc@{}}
\toprule
\multirow{3}{*}{Method} &
  \multicolumn{10}{c}{Biomarkers} &
  \multicolumn{3}{c}{Metrics} \\ \cmidrule(l){2-14} 
 &
  \multicolumn{2}{c}{IRF} &
  \multicolumn{2}{c}{DRT/ME} &
  \multicolumn{2}{c}{IRHRF} &
  \multicolumn{2}{c}{FAVF} &
  \multicolumn{2}{c}{PAVF} &
  \multicolumn{1}{c}{\multirow{2}{*}{AUROC}} &
  \multicolumn{1}{c}{\multirow{2}{*}{\begin{tabular}[c]{@{}c@{}}Average \\ Specificity\end{tabular}}} &
  \multirow{2}{*}{\begin{tabular}[c]{@{}c@{}}Average \\ Sensitivity\end{tabular}} \\
 &
  \multicolumn{1}{c}{Accuracy} &
  \multicolumn{1}{c}{F1-Score} &
  \multicolumn{1}{c}{Accuracy} &
  \multicolumn{1}{c}{F1-Score} &
  \multicolumn{1}{c}{Accuracy} &
  \multicolumn{1}{c}{F1-Score} &
  \multicolumn{1}{c}{Accuracy} &
  \multicolumn{1}{c}{F1-Score} &
  \multicolumn{1}{c}{Accuracy} &
  \multicolumn{1}{c}{F1-Score} &
  \multicolumn{1}{c}{} &
  \multicolumn{1}{c}{} &
   \\ \hline \midrule
\multicolumn{1}{c}{PCL \cite{li2020prototypical}} &
  \textbf{76.50\%} &
  \multicolumn{1}{c}{0.717} &
  80.11\% &
  \multicolumn{1}{c}{0.761} &
  59.10\% &
  \multicolumn{1}{c}{0.683} &
  76.30\% &
  \multicolumn{1}{c}{0.773} &
  51.40\% &
  \multicolumn{1}{c}{0.165} &
  \multicolumn{1}{c}{0.767} &
  \multicolumn{1}{c}{0.741} &
  0.604 \\
\multicolumn{1}{c}{SimCLR \cite{chen2020simple}} &
  75.13\% &
  \multicolumn{1}{c}{0.716} &
  80.61\% &
  \multicolumn{1}{c}{0.772} &
  59.03\% &
  \multicolumn{1}{c}{0.675} &
  75.43\% &
  \multicolumn{1}{c}{0.761} &
  52.69\% &
  \multicolumn{1}{c}{0.249} &
  \multicolumn{1}{c}{0.754} &
  \multicolumn{1}{c}{0.747} &
  0.614 \\
\multicolumn{1}{c}{Moco V2 \cite{chen2020improved}} &
  76.00\% &
  \multicolumn{1}{c}{\textbf{0.720}} &
  82.24\% &
  \multicolumn{1}{c}{0.793} &
  59.60\% &
  \multicolumn{1}{c}{0.692} &
  75.00\% &
  \multicolumn{1}{c}{0.784} &
  52.69\% &
  \multicolumn{1}{c}{0.211} &
  \multicolumn{1}{c}{0.770} &
  \multicolumn{1}{c}{0.762} &
  0.651 \\ \hline \midrule
\multicolumn{1}{c}{Eye ID} &
  72.63\% &
  \multicolumn{1}{c}{0.674} &
  80.20\% &
  \multicolumn{1}{c}{0.778} &
  58.00\% &
  \multicolumn{1}{c}{0.674} &
  74.93\% &
  \multicolumn{1}{c}{0.725} &
  \textbf{65.56\%} &
  \multicolumn{1}{c}{\textbf{0.588}} &
  \multicolumn{1}{c}{0.767} &
  \multicolumn{1}{c}{\textbf{0.776}} &
  0.656 \\
\multicolumn{1}{c}{CST} &
  75.53\% &
  \multicolumn{1}{c}{0.720} &
  \textbf{83.06\%} &
  \multicolumn{1}{c}{\textbf{0.811}} &
  \textbf{64.30\%} &
  \multicolumn{1}{c}{\textbf{0.703}} &
  76.13\% &
  \multicolumn{1}{c}{0.766} &
  62.16\% &
  \multicolumn{1}{c}{0.509} &
  \multicolumn{1}{c}{\textbf{0.790}} &
  \multicolumn{1}{c}{0.772} &
  \textbf{0.675} \\
\multicolumn{1}{c}{BCVA} &
  74.03\% &
  \multicolumn{1}{c}{0.701} &
  80.27\% &
  \multicolumn{1}{c}{0.770} &
  58.8\% &
  \multicolumn{1}{c}{0.672} &
  \textbf{77.63\%} &
  \multicolumn{1}{c}{\textbf{0.785}} &
  58.06\% &
  \multicolumn{1}{c}{0.418} &
  \multicolumn{1}{c}{0.776} &
  \multicolumn{1}{c}{0.713} &
  0.645 \\ \bottomrule
\end{tabular}%
}
\end{table}
\section*{Conclusion}

\paragraph{Broader Impacts}

From a medical perspective, our paper shows that there are
ways to utilize correlations that exist between measured clinical labels
and their associated biomarker structures within images. This work potentially inspires medical research
into other domains and clinical settings where questions exist as
to how to effectively utilize relationships that exist within the data
available. This is especially relevant in contexts where access to one modality is easier than another.

\paragraph{Discussion}

In this work, we investigate how the usage of a supervised
contrastive loss on clinical data can be used to effectively train a
model for the task of biomarker classification. We show how the
method performs across different combinations of clinical labels. We conclude that
the usage of the clinical labels is a more effective way to leverage
the correlations that exist within unlabeled data over traditional contrastive learning algorithms. We prove this through extensive
experimentation on biomarkers of varying granularity within OCT
scans.

\newpage

\bibliographystyle{IEEEbib}
\bibliography{ref}

\begin{thebibliography}{10}

\bibitem{le2020contrastive}
Phuc~H Le-Khac, Graham Healy, and Alan~F Smeaton,
\newblock ``Contrastive representation learning: A framework and review,''
\newblock {\em IEEE Access}, 2020.

\bibitem{chen2020simple}
Ting Chen, Simon Kornblith, Mohammad Norouzi, and Geoffrey Hinton,
\newblock ``A simple framework for contrastive learning of visual
  representations,''
\newblock in {\em International conference on machine learning}. PMLR, 2020,
  pp. 1597--1607.

\bibitem{strimbu2010biomarkers}
Kyle Strimbu and Jorge~A Tavel,
\newblock ``What are biomarkers?,''
\newblock {\em Current Opinion in HIV and AIDS}, vol. 5, no. 6, pp. 463, 2010.

\bibitem{mcdonald2015effects}
Robert~J McDonald, Kara~M Schwartz, Laurence~J Eckel, Felix~E Diehn,
  Christopher~H Hunt, Brian~J Bartholmai, Bradley~J Erickson, and David~F
  Kallmes,
\newblock ``The effects of changes in utilization and technological
  advancements of cross-sectional imaging on radiologist workload,''
\newblock {\em Academic radiology}, vol. 22, no. 9, pp. 1191--1198, 2015.

\bibitem{prabhushankar2022olives}
Mohit Prabhushankar, Kiran Kokilepersaud, Yash-yee Logan, Stephanie
  Trejo~Corona, Ghassan AlRegib, and Charles Wykoff,
\newblock ``Olives dataset: Ophthalmic labels for investigating visual eye
  semantics,''
\newblock in {\em Proceedings of the Neural Information Processing Systems
  Track on Datasets and Benchmarks 2 (NeurIPS Datasets and Benchmarks 2022)},
  2022.

\bibitem{hannouche2012correlation}
Rosana~Zacarias Hannouche, Marcos Pereira~de {\'A}vila, David Leonardo~Cruvinel
  Isaac, Alan~Ricardo Rassi, et~al.,
\newblock ``Correlation between central subfield thickness, visual acuity and
  structural changes in diabetic macular edema,''
\newblock {\em Arquivos brasileiros de oftalmologia}, vol. 75, no. 3, pp.
  183--187, 2012.

\bibitem{sun2014disorganization}
Jennifer~K Sun, Michael~M Lin, Jan Lammer, Sonja Prager, Rutuparna Sarangi,
  Paolo~S Silva, and Lloyd~Paul Aiello,
\newblock ``Disorganization of the retinal inner layers as a predictor of
  visual acuity in eyes with center-involved diabetic macular edema,''
\newblock {\em JAMA ophthalmology}, vol. 132, no. 11, pp. 1309--1316, 2014.

\bibitem{murakami2011association}
Tomoaki Murakami, Kazuaki Nishijima, Atsushi Sakamoto, Masafumi Ota, Takahiro
  Horii, and Nagahisa Yoshimura,
\newblock ``Association of pathomorphology, photoreceptor status, and retinal
  thickness with visual acuity in diabetic retinopathy,''
\newblock {\em American journal of ophthalmology}, vol. 151, no. 2, pp.
  310--317, 2011.

\bibitem{kashani2010retinal}
Amir~H Kashani, Ingrid~E Zimmer-Galler, Syed~Mahmood Shah, Laurie Dustin,
  Diana~V Do, Dean Eliott, Julia~A Haller, and Quan~Dong Nguyen,
\newblock ``Retinal thickness analysis by race, gender, and age using stratus
  oct,''
\newblock {\em American journal of ophthalmology}, vol. 149, no. 3, pp.
  496--502, 2010.

\bibitem{arora2019theoretical}
Sanjeev Arora, Hrishikesh Khandeparkar, Mikhail Khodak, Orestis Plevrakis, and
  Nikunj Saunshi,
\newblock ``A theoretical analysis of contrastive unsupervised representation
  learning,''
\newblock {\em arXiv preprint arXiv:1902.09229}, 2019.

\bibitem{khosla2020supervised}
Prannay Khosla, Piotr Teterwak, Chen Wang, Aaron Sarna, Yonglong Tian, Phillip
  Isola, Aaron Maschinot, Ce~Liu, and Dilip Krishnan,
\newblock ``Supervised contrastive learning,''
\newblock {\em arXiv preprint arXiv:2004.11362}, 2020.

\bibitem{chen2020improved}
Xinlei Chen, Haoqi Fan, Ross Girshick, and Kaiming He,
\newblock ``Improved baselines with momentum contrastive learning,''
\newblock {\em arXiv preprint arXiv:2003.04297}, 2020.

\bibitem{li2020prototypical}
Junnan Li, Pan Zhou, Caiming Xiong, and Steven~CH Hoi,
\newblock ``Prototypical contrastive learning of unsupervised
  representations,''
\newblock {\em arXiv preprint arXiv:2005.04966}, 2020.

\bibitem{he2016deep}
Kaiming He, Xiangyu Zhang, Shaoqing Ren, and Jian Sun,
\newblock ``Deep residual learning for image recognition,''
\newblock in {\em Proceedings of the IEEE conference on computer vision and
  pattern recognition}, 2016, pp. 770--778.

\bibitem{cole2021does}
Elijah Cole, Xuan Yang, Kimberly Wilber, Oisin Mac~Aodha, and Serge Belongie,
\newblock ``When does contrastive visual representation learning work?,''
\newblock {\em arXiv preprint arXiv:2105.05837}, 2021.

\bibitem{markan2020novel}
Ashish Markan, Aniruddha Agarwal, Atul Arora, Krinjeela Bazgain, Vipin Rana,
  and Vishali Gupta,
\newblock ``Novel imaging biomarkers in diabetic retinopathy and diabetic
  macular edema,''
\newblock {\em Therapeutic Advances in Ophthalmology}, vol. 12, pp.
  2515841420950513, 2020.

\bibitem{kermany2018labeled}
Daniel Kermany, Kang Zhang, Michael Goldbaum, et~al.,
\newblock ``Labeled optical coherence tomography (oct) and chest x-ray images
  for classification,''
\newblock {\em Mendeley data}, vol. 2, no. 2, 2018.

\bibitem{farsiu2014quantitative}
Sina Farsiu, Stephanie~J Chiu, Rachelle~V O'Connell, Francisco~A Folgar, Eric
  Yuan, Joseph~A Izatt, Cynthia~A Toth, Age-Related Eye Disease Study 2
  Ancillary Spectral Domain Optical Coherence Tomography~Study Group, et~al.,
\newblock ``Quantitative classification of eyes with and without intermediate
  age-related macular degeneration using optical coherence tomography,''
\newblock {\em Ophthalmology}, vol. 121, no. 1, pp. 162--172, 2014.

\bibitem{deng2009imagenet}
Jia Deng, Wei Dong, Richard Socher, Li-Jia Li, Kai Li, and Li~Fei-Fei,
\newblock ``Imagenet: A large-scale hierarchical image database,''
\newblock in {\em 2009 IEEE conference on computer vision and pattern
  recognition}. Ieee, 2009, pp. 248--255.

\bibitem{he2020momentum}
Kaiming He, Haoqi Fan, Yuxin Wu, Saining Xie, and Ross Girshick,
\newblock ``Momentum contrast for unsupervised visual representation
  learning,''
\newblock in {\em Proceedings of the IEEE/CVF Conference on Computer Vision and
  Pattern Recognition}, 2020, pp. 9729--9738.

\bibitem{caron2020unsupervised}
Mathilde Caron, Ishan Misra, Julien Mairal, Priya Goyal, Piotr Bojanowski, and
  Armand Joulin,
\newblock ``Unsupervised learning of visual features by contrasting cluster
  assignments,''
\newblock {\em arXiv preprint arXiv:2006.09882}, 2020.

\bibitem{grill2020bootstrap}
Jean-Bastien Grill, Florian Strub, Florent Altch{\'e}, Corentin Tallec,
  Pierre~H Richemond, Elena Buchatskaya, Carl Doersch, Bernardo~Avila Pires,
  Zhaohan~Daniel Guo, Mohammad~Gheshlaghi Azar, et~al.,
\newblock ``Bootstrap your own latent: A new approach to self-supervised
  learning,''
\newblock {\em arXiv preprint arXiv:2006.07733}, 2020.

\bibitem{azizi2021big}
Shekoofeh Azizi, Basil Mustafa, Fiona Ryan, Zachary Beaver, Jan Freyberg,
  Jonathan Deaton, Aaron Loh, Alan Karthikesalingam, Simon Kornblith, Ting
  Chen, et~al.,
\newblock ``Big self-supervised models advance medical image classification,''
\newblock {\em arXiv preprint arXiv:2101.05224}, 2021.

\bibitem{chen2021disease}
Yen-Pin Chen, Yuan-Hsun Lo, Feipei Lai, and Chien-Hua Huang,
\newblock ``Disease concept-embedding based on the self-supervised method for
  medical information extraction from electronic health records and disease
  retrieval: Algorithm development and validation study,''
\newblock {\em Journal of Medical Internet Research}, vol. 23, no. 1, pp.
  e25113, 2021.

\bibitem{vu2021medaug}
Yen Nhi~Truong Vu, Richard Wang, Niranjan Balachandar, Can Liu, Andrew~Y Ng,
  and Pranav Rajpurkar,
\newblock ``Medaug: Contrastive learning leveraging patient metadata improves
  representations for chest x-ray interpretation,''
\newblock {\em arXiv preprint arXiv:2102.10663}, 2021.

\end{thebibliography}

\newpage
\section*{Checklist}

\begin{enumerate}

\item For all authors...
\begin{enumerate}
  \item Do the main claims made in the abstract and introduction accurately reflect the paper's contributions and scope?
    \answerYes{}
  \item Did you describe the limitations of your work?
    \answerYes{} See Appendix \ref{app:limits}.
  \item Did you discuss any potential negative societal impacts of your work?
    \answerYes{} See Appendix \ref{app:limits}.
  \item Have you read the ethics review guidelines and ensured that your paper conforms to them?
    \answerYes{}
\end{enumerate}

\item If you are including theoretical results...
\begin{enumerate}
  \item Did you state the full set of assumptions of all theoretical results?
    \answerYes{} See Theoretical Discussion Section
        \item Did you include complete proofs of all theoretical results?
    \answerNA{} Theory from paper \cite{arora2019theoretical} was discussed and showed connections to this paper, no explicit proof was necessary.
\end{enumerate}

\item If you ran experiments...
\begin{enumerate}
  \item Did you include the code, data, and instructions needed to reproduce the main experimental results (either in the supplemental material or as a URL)?
    \answerYes{} See Appendix \ref{app: links}
  \item Did you specify all the training details (e.g., data splits, hyperparameters, how they were chosen)?
    \answerYes{} See Appendix \ref{app:contrastive}
        \item Did you report error bars (e.g., with respect to the random seed after running experiments multiple times)?
    \answerYes{} See Appendix \ref{app:results}
        \item Did you include the total amount of compute and the type of resources used (e.g., type of GPUs, internal cluster, or cloud provider)?
    \answerYes{} See Appendix \ref{app:resources}
\end{enumerate}

\item If you are using existing assets (e.g., code, data, models) or curating/releasing new assets...
\begin{enumerate}
  \item If your work uses existing assets, did you cite the creators?
    \answerYes{} See Appendix \ref{app:dataset}.
  \item Did you mention the license of the assets?
    \answerYes{} See Appendix \ref{app: links}.
  \item Did you include any new assets either in the supplemental material or as a URL?
    \answerNA{}
  \item Did you discuss whether and how consent was obtained from people whose data you're using/curating?
    \answerYes{} The dataset is open source.
  \item Did you discuss whether the data you are using/curating contains personally identifiable information or offensive content?
    \answerYes{} The open source dataset discusses this point in their paper. There is no personally identifiable information within the dataset.
\end{enumerate}

\item If you used crowdsourcing or conducted research with human subjects...
\begin{enumerate}
  \item Did you include the full text of instructions given to participants and screenshots, if applicable?
    \answerNA{}
  \item Did you describe any potential participant risks, with links to Institutional Review Board (IRB) approvals, if applicable?
    \answerNA{}
  \item Did you include the estimated hourly wage paid to participants and the total amount spent on participant compensation?
    \answerNA{}
\end{enumerate}

\end{enumerate}

\newpage

\appendix

\section{Dataset Description}
\subsection{Dataset Details}
\label{app:dataset}

\begin{figure}[t!]
\centering
\includegraphics[scale = .3]{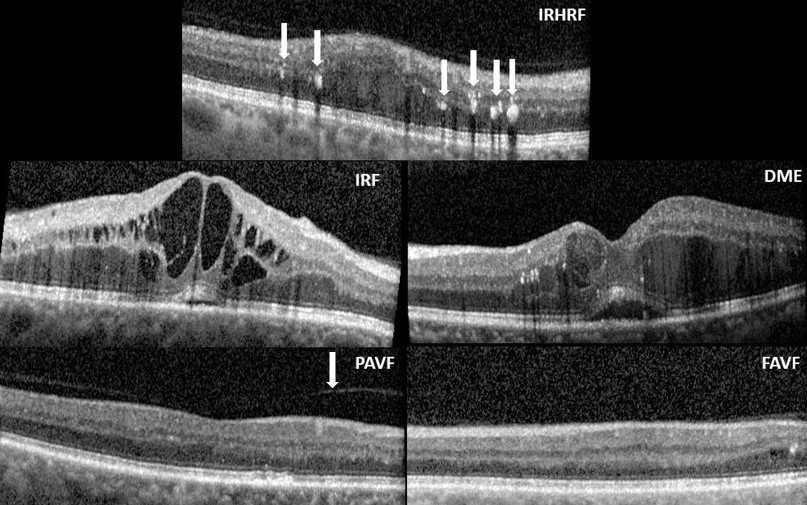}
\caption{ Cross-sectional images of graded biomarkers. Intra-Retinal Hyper-Reflective Foci (IRHRF),
indicated by the six white arrows, are areas of hyperreflectivity in the intraretinal layers with or
without shadowing of the more posterior retinal layers. Intra-Retinal Fluid (IRF) encompasses
the cystic areas of hyporeflectivity. Diabetic Macular Edema (DME) is the apparent swelling and
elevation of the macula due to the presence of fluid. A Partially Attached Vitreous Face (PAVF), with
an arrow indicating the point of attachment and a Fully Attached Vitreous Face (FAVF). A discussion
of these biomarkers can be found at \cite{markan2020novel}. }\label{fig:biomarkers}
\end{figure}

\begin{figure}[t!]
\centering
\includegraphics[scale = .4]{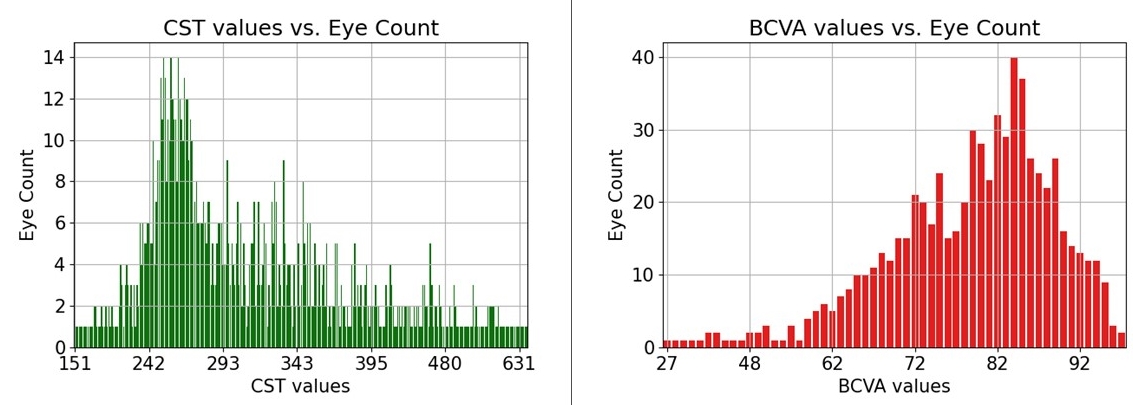}
\caption{Distribution of CST and BCVA labels in OLIVES dataset based on number of eyes associated
with each clinical value. }\label{fig:distribution}
\end{figure}

The dataset is used in this study is the \texttt{OLIVES} dataset \cite{prabhushankar2022olives}. Full details surrounding the dataset can be understood from the paper, but we wish to highlight specific figures from the paper to understand the studies performed in this workshop. Table \ref{tab:olives} provides a full description of all the labels available within the dataset. It specifies the exact biomarkers and clinical labels obtained for every image as well as how they were collected. Figure \ref{fig:biomarkers} shows examples of the biomarkers that are detected in the studies shown in this paper. Furthermore, a complete analysis of the distributions of clinical values present in the study can be found at Figure \ref{fig:distribution}.

\begin{table}[h!]
\small
\caption{High-level overview of the \texttt{OLIVES} Dataset. The modality column details the type of data. The columns "Per Visit" and "Per Eye" indicate the amount of data in each modality on a respective visit or eye. $N_P$ is the number of visits that a patient $P$ takes to the clinic. The statistics across all eyes across all visits are shown in the Total Statistics column. Biomarkers are binary values, clinical labels are integers, fundus are 2D images, and OCT are 3D slices.}\label{tab:olives}
\begin{tabular}{@{}ccccc@{}}
\toprule
\multicolumn{5}{c}{OLIVES Dataset Summary}                        \\ \midrule
Modality & Per Visit & Per Eye & Total Statistics      & Overview \\ \midrule
\multirow{5}{*}{\begin{tabular}[c]{@{}c@{}}\\OCT\\ \\ Fundus\\ \\ Clinical\\ \\ Biomarker\end{tabular}} &
  \multirow{5}{*}{\begin{tabular}[c]{@{}c@{}}\\49\\ \\ 1\\ \\ 4\\ \\ 16\end{tabular}} &
  \multirow{5}{*}{\begin{tabular}[c]{@{}c@{}}\\$N_P$*49\\ \\ $N_P$\\ \\ $N_P$*4\\ \\ 1568\end{tabular}} &
  \multicolumn{1}{c|}{\multirow{5}{*}{\begin{tabular}[c]{@{}c@{}}\\78189\\ \\ 1268\\ \\ 5072\\ \\ 150528\end{tabular}}} &
  \multirow{5}{*}{\begin{tabular}[c]{@{}c@{}}\textbf{General:}\\ 96 Eyes, Visits every 4-16 weeks, \\ Average 16 visits and 7 injections/patient\\ \textbf{Clinical Labels obtained every visit:}\\ BCVA, CST, Patient ID, Eye ID\\ \textbf{Biomarkers labeled:}\\ IRHRF, FAVF, IRF, DRT/ME PAVF, VD, \\ Preretinal Tissue, EZ Disruption, IR Hemmorhages, \\ SRF, VMT, Atrophy, SHRM, RPE Disruption,\\ Serous PED\end{tabular}} \\
         &           &         & \multicolumn{1}{c|}{} &          \\
         &           &         & \multicolumn{1}{c|}{} &          \\
         &           &         & \multicolumn{1}{c|}{} &          \\
         &           &         & \multicolumn{1}{c|}{} &          \\
         &           &         & \multicolumn{1}{c|}{} &          \\
         &           &         & \multicolumn{1}{c|}{} &         \\
         &           &         & \multicolumn{1}{c|}{} &          \\
         &           &         & \multicolumn{1}{c|}{} &         \\
         &           &         & \multicolumn{1}{c|}{} &         
         \\\bottomrule
\label{tab:olives}
\end{tabular}
\end{table}

\subsection{Limitations}
\label{app:limits}

The primary limitations of our study is the number of clinical labels we investigated as well as the fact the dataset comes from people who attended a single clinic. Having a wider range of clinical labels would allow us to explore how the wide range of available medical data transfers in terms of learning representations for other related tasks. Additionally, the fact the dataset comes from a single location limits how well the model can generalize towards other populations of patients. This is an inherent limitation of any medical dataset and could potentially be accounted for through supplementing analysis with other datasets such as Kermany \cite{kermany2018labeled} and Farisu \cite{farsiu2014quantitative}. This potential lack of generalization towards other sub-populations is problematic if not properly accounted for when applying algorithms to real-world settings. Without this accountability, this could have a negative societal impact. 
\section{Additional Results and Training Details}
\subsection{Standard Deviation}
\label{app:results}

The standard deviations for 3 different runs of the algorithm in the task of individual biomarker detection and average AUROC are shown in Table \ref{tab:main_std}.

\begin{table*}[h!]
\centering
\scriptsize
\begin{tabular}{ccccccc}
\toprule
\multirow{2}{*}{Method} & \multicolumn{6}{c}{Biomarkers}  \\ \cmidrule(l){2-7} 
 & \multicolumn{1}{c}{IRF} & \multicolumn{1}{c}{DME} & \multicolumn{1}{c}{IRHRF} & \multicolumn{1}{c}{FAVF} & \multicolumn{1}{c}{PAVF} & \multicolumn{1}{c}{AUROC} \\
 \hline \midrule
PCL  \cite{li2020prototypical}              & \textbf{76.50}\% $\pm$ .513   & 80.11\% $\pm$ .335  & 59.1\% $\pm$ 1.03 & 76.30\% $\pm$ .378 & 51.40\% $\pm$ .556  & .767 $\pm$ .0017  \\

SimCLR \cite{chen2020simple}                    & 75.13\% $\pm$ .529  & 80.61\% $\pm$ .837  & 59.03\% $\pm$ 2.54  & 75.43\% $\pm$ .378  & 52.69\% $\pm$ 2.68 & .754 $\pm$ .0017\\
Moco v2 \cite{chen2020improved}                   & 76.00\%  $\pm$ .305     &  82.24\%  $\pm$ 1.38      & 59.6\%  $\pm$ .702      &  75.00\%   $\pm$ .608    &  52.69\% $\pm$ .472   & .770 $\pm$ .0035 \\ \hline \midrule

Eye ID                     & 72.63\% $\pm$ .264  & 80.2\%  $\pm$ .384  & 58\% $\pm$ 2.56   & 74.93\%  $\pm$ 1.36  & \textbf{65.56}\% $\pm$ .200  & .767 $\pm$ .0005 \\
CST                        & 75.53\%  $\pm$ .608  & \textbf{83.06}\% $\pm$ .213    & \textbf{64.3}\% $\pm$ 2.57  & 76.13\% $\pm$ .264   & 62.16\% $\pm$ 1.47  & .790 $\pm$ .0006\\
BCVA                       & 74.03\% $\pm$ .351  & 80.27\% $\pm$ .853  & 58.8\% $\pm$ 1.82  & \textbf{77.63}\% $\pm$ .305  & 58.06\% $\pm$ 1.27  & .776 $\pm$ .0017  \\

 \bottomrule
\end{tabular}
\caption{We show the performance of supervised contrastive training on the OLIVES dataset. In this table we explicitly show the standard deviation for the average across three runs for both accuracy and AUROC.}
\label{tab:main_std}
\end{table*}
\label{app:contrastive}
\subsection{Additional Training Details}
Given an input batch of data, $(x_{k}$, and clinical labels, $(y_{k}$ we obtain the set $(x_{k},y_{k})_{k=1,...,N}$. We perform augmentations on the batch twice in order to get two copies of the original batch with $2N$ images and clinical labels. These augmentations are random resize crop to a size of 224, random horizontal flips, random color jitter, and data normalization. These are sensible from a medical perspective because they don't disrupt the general structure of the retina.
This process produces a larger set $(x_{l},y_{l})_{l=1,...,2N}$ that consists of two versions of each image that differ only due to the random nature of the applied augmentation. Thus, for every image $x_{k}$ and clinical label $y_{k}$ there exists two views of the image $x_{2k}$ and $x_{2k-1}$ and two copies of the clinical labels that are equivalent to each other: $y_{2k-1} = y_{2k} = y_{k}$.

\begin{figure}[t!]
\centering
\includegraphics[scale = .3]{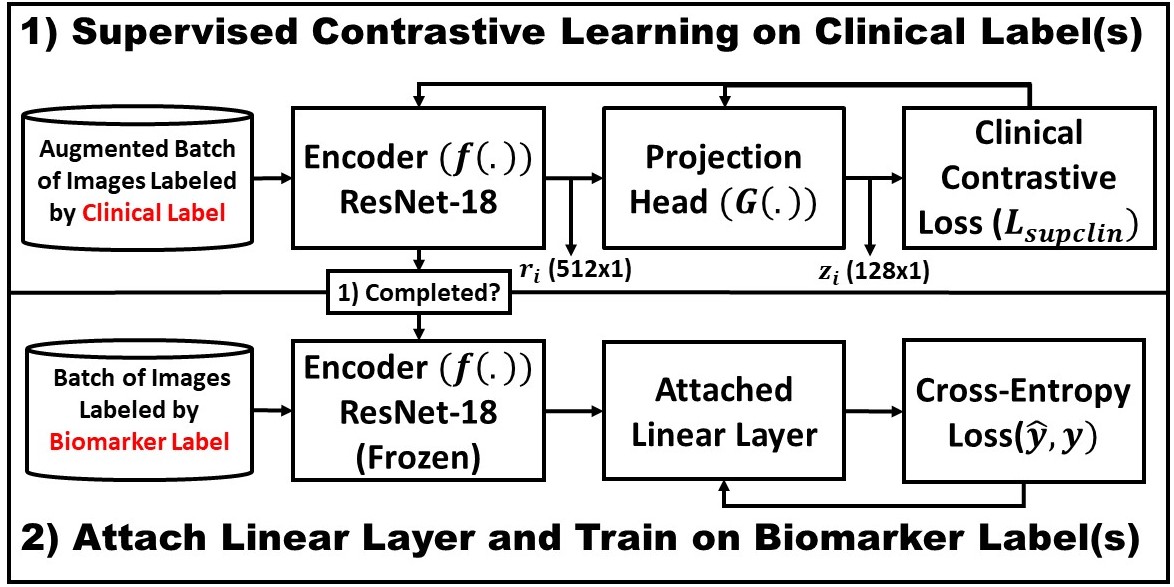}
\caption{Overview of supervised contrastive learning and linear fine-
tuning steps. 1) Supervised Contrastive Loss on clinical
labels. 2) Attach linear layer and
train on labeled biomarker data. }\label{fig:methodology}
\end{figure}

From this point, we perform the first step in Figure \ref{fig:methodology}, where supervised contrastive learning is performed on the identified clinical label. The clinically labeled augmented batch is forward-propagated through an encoder network $f(\cdot)$ that we set to be the ResNet-18 architecture \cite{he2016deep}. This results in a 512-dimensional vector $r_{i}$ that is sent through a projection network $G(\cdot)$, which further compresses the representation to a 128-dimensional embedding vector $z_{i}$. $G(\cdot)$ is chosen to be a multi-layer perceptron network with a single hidden layer. This projection network is utilized only to reduce the dimensionality of the embedding before computing the loss and is discarded after training. A supervised contrastive loss \cite{khosla2020supervised} is performed on the output of the projection network in order to train the encoder network. In this case, embeddings with the same clinical label are enforced to be projected closer to each other while embeddings with differing clinical labels are projected away from each other.  Our introduced clinical supervised contrastive loss process can be understood by:
$$
    L_{supconclin} = \sum_{i\in{I}} \frac{-1}{|C(i)|}\sum_{c\in{C(i)}}log\frac{exp(z_{i}\cdot z_{c}/\tau)}{\sum_{a\in{A(i)}}exp(z_{i}\cdot z_{a}/\tau)}
$$
where $i$ is the index for the image of interest $x_{i}$.
All positives $c$ for image $x_{i}$ are obtained from the set $C(i)$ and all positive and negative instances $a$ are obtained from the set $A(i)$. Every element $c$ of $C(i)$ represents all other images in the batch with the same clinical label $c$ as the image of interest $x_{i}$. Additionally, $z_{i}$ is the embedding for the image of interest, $z_{c}$ represents the embedding for the clinical positives, and $z_{a}$ represents the embeddings for all positive and negative instances in the set $A(i)$. $\tau$ is a temperature scaling parameter that is set to .07 for all experiments.  The loss function operates in the embedding space where the goal is to maximize the cosine similarity between embedding $z_{i}$ and its set of clinical positives $z_{c}$. It should be explicitly stated that the set $C(i)$ can represent any clinical label of interest.

After training the encoder with  clinical supervised contrastive loss, we move to the second step in Figure \ref{fig:methodology} where the weights of the encoder are frozen and a linear layer is appended to the output of the encoder. This setup is trained on the available biomarker data after choosing the biomarker we wish to train for. This linear layer is trained using cross-entropy loss to distinguish between the presence or absence of the biomarker of interest in the OCT scan. In this way, we leverage knowledge learnt from training on clinical labels to improve performance on classifying biomarkers. The previously trained encoder with the supervised contrastive loss on the clinical label from step 1 produces the representation for the input and this representation is fine-tuned with the linear layer to distinguish whether or not the biomarker of interest is present. 

Care was taken to ensure that all aspects of the experiments
remained the same whether training was done via supervised or
self-supervised contrastive learning on the encoder or cross-entropy
training on the attached linear classifier. The encoder utilized was
kept as a ResNet-18 architecture. The applied augmentations are
random resize crop to a size of 224, random horizontal flips,
random color jitter, and data normalization to the mean and standard
deviation of the respective dataset. The batch size was set at 64.
Training was performed for 25 epochs in every setting. A stochastic
gradient descent optimizer was used with a learning rate of 1e-3 and
momentum of .9.

\subsection{Training of Comparison Architectures}

The training for SimCLR, Moco v2, and PCL was performed using the same setup described for the supervised contrastive learning experiments described in the previous section. This means the same ResNet-18 architecture and MLP projection head was utilized. It should be noted we made appropriate changes to ResNet-18 to fit the constraints of our dataset i.e. single channel images as well as removal of an early max-pooling layer. All models were trained for 25 epochs with a batch size of 64. The applied augmentations were  random resize crop to a size of 224, random horizontal flips, random color jitter, and data normalization to the mean and standard deviation of the respective dataset. A stochastic gradient descent optimizer with a learning rate of 1e-3 and momentum of .9 was used during pre-training of the backbone model as well as the appended linear layer during the biomarker classification step. Links to the code for these strategies can be found in Appendix \ref{app: links}.

Additional details exist for Moco v2 and PCL in regards to certain hyperparameters introduced within their respective methodologies. For Moco v2, we used the same default hyper-parameters as introduced in the original paper. This includes a queue size of 65536, a moco momentum update term of .999, and a softmax temperature of .07. For PCL, further changes had to be made due to the clustering nature of PCL. In this case, the provided hyperparameters of PCL were based on the size of the Imagenet \cite{deng2009imagenet} dataset which has in excess of 1 million images. However, our training set had 60,000 images, which means that the default clustering values were inappropriate. The changes we made to fit our constraints included a queue size of 64, the number of clusters set to [100, 200, 300], the moco momentum term set to .999, and the softmax temperature scaling set to .07. 

\section{Code and Resources}
\subsection{Links to Access Dataset and Code}
\label{app: links}
We provide open access to the dataset. The images and labels found in the OLIVES dataset are present at:\\
 \href{https://doi.org/10.5281/zenodo.6622145}{Dataset Access}\\
The code for the benchmarks provided in the paper are accessible at the following link:  \\
 \href{https://github.com/olivesgatech/OLIVES_Dataset}{Code Access}

 \href{https://github.com/salesforce/PCL}{PCL}
 
The code is associated with an MIT License. The DOI of the dataset is 10.5281/zenodo.6622145.
The associated license with the dataset is a Creative Commons International 4 license.

 The code for the comparison architectures can be found at:

 \href{https://github.com/HobbitLong/SupContrast}{SimCLR}

  \href{https://github.com/facebookresearch/moco}{Moco v2}

 \subsection{Computing Resources}
 \label{app:resources}
 All experiments were run on PCs with  NVIDIA GeForce GTX TITAN X 12 GB GPUs.
\section{Related Works}

Modern contrastive learning approaches such as \cite{chen2020simple,he2020momentum,caron2020unsupervised, grill2020bootstrap} all generate positive pairs of images through various types of data augmentations such as random cropping, multi-cropping, and different types of blurs and color jitters. A classifier can then be trained on top of these learned representations while requiring fewer labels for satisfactory performance.
  Recent work has explored the idea of using medically consistent meta-data as a means of finding positive pairs of images alongside augmentations for a contrastive loss function. \cite{azizi2021big} showed that using images from the same medical pathology as well as augmentations for positive image pairs could improve representations beyond standard self-supervision. \cite{chen2021disease} demonstrated utilizing contrastive learning with a transformer can learn embeddings for electronic health records that can correlate with various disease concepts. \cite{vu2021medaug} investigated choosing positive pairs from images that exist from the same patient, clinical study, and laterality. These works demonstrate the potential of utilizing clinical data within a contrastive learning framework. However, these methods were tried on limited clinical data settings, such as choosing images from the same patient or position relative to other tissues. Our work builds on these ideas by explicitly using measured clinical labels from an eye-disease setting as its own label for training a model. By doing this, we present a comprehensive assessment of what kinds of clinical data can possibly be used as a means of choosing positive instances from the perspective of OCT scans and expand the scope of how clinical data can be utilized.

\end{document}